\pdfoutput=1

\documentclass[11pt]{article}

\usepackage[final]{acl}

\usepackage{times}
\usepackage{latexsym}

\usepackage[T1]{fontenc}

\usepackage[utf8]{inputenc}

\usepackage{microtype}

\usepackage{amsfonts}
\usepackage{amssymb}

\usepackage{pifont}
\usepackage{bm}
\usepackage{arydshln}
\usepackage{amsmath}
\usepackage{xcolor}
\usepackage{colortbl}
\usepackage{tcolorbox}
\usepackage{mathtools}
\usepackage{adjustbox}
\usepackage{multirow}
\usepackage{paralist}
\usepackage{makecell}
\usepackage{booktabs}
\usepackage{nicefrac}
\usepackage{enumerate}
\usepackage{bbding} 
\usepackage{wasysym}
\usepackage{inconsolata}
\usepackage{enumitem}

\usepackage{todonotes}
\definecolor{chong-color}{rgb}{0.858, 0.188, 0.478}

\newcommand{\greentick}{\textcolor{green!70!black}{\ding{52}}}
\newcommand{\redcross}{\textcolor{red}{\ding{56}}}

\newcommand\datasetname{\textsc{GlobeSumm}}
\newcommand\taskname{\textsc{McMs}}

\newtcolorbox{promptbox}[2][]{
	width=\columnwidth,
	colback = gray!8, 
	colframe = gray!8, 
	boxsep=0pt,left=10pt,right=10pt,top=0pt,bottom=0pt,
	fontupper=\linespread{0.9}\selectfont,
	title=#2,#1,
        fontupper=\small}

\definecolor{color1}{RGB}{240,230,140}
\definecolor{color2}{RGB}{197,217,197}
\definecolor{color3}{RGB}{225,179,191}
\definecolor{color4}{RGB}{176,224,230}
\definecolor{where}{RGB}{250,128,114}
\definecolor{strategy}{RGB}{176,224,230}

\definecolor{forestgreen}{RGB}{34, 139, 34}
\definecolor{firebrick}{RGB}{178, 34, 34}

\newcommand{\uparr}{$\color{firebrick}{\bm{\uparrow}}$}
\newcommand{\downarr}{$\color{forestgreen}{\bm{\downarrow}}$}

\title{{\includegraphics[width=25pt]{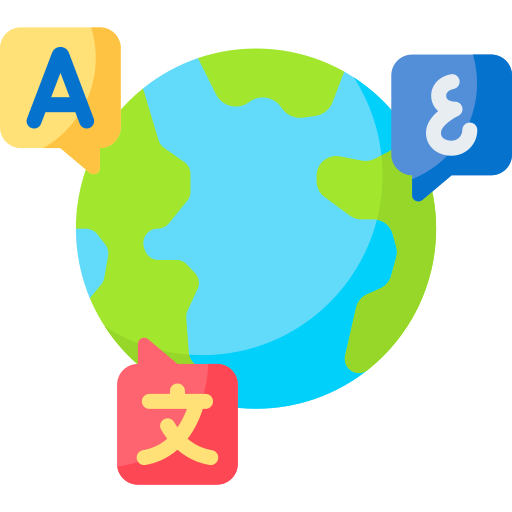}~\datasetname}: A Challenging Benchmark Towards Unifying Multi-lingual, Cross-lingual and Multi-document News Summarization}

\author{
  Yangfan Ye$^{1}$,
  Xiachong Feng$^{2}$,
  Xiaocheng Feng$^{1,3}$\thanks{Corresponding Author},
  Weitao Ma$^{1}$,
  Libo Qin$^{4}$ \\
  \textbf{Dongliang Xu$^{5}$,
  Qing Yang$^{5}$,
  Hongtao Liu$^{5}$,
  Bing Qin$^{1,3}$} \\
  $^{1}$Harbin Institute of Technology \quad
  $^{2}$The University of Hong Kong \quad
  $^{3}$Peng Cheng Laboratory \\
  $^{4}$Central South University \quad \quad
  $^{5}$Du Xiaoman Financial, Beijing \\
  \texttt{\{yfye,xcfeng,wtma,qinb\}@ir.hit.edu.cn} \quad \texttt{fengxc@hku.hk} \\
  \texttt{lbqin@csu.edu.cn} \quad \texttt{\{xudongliang,yangqing,liuhongtao01\}@duxiaoman.com}
}

\begin{document}
\maketitle
\begin{abstract}
News summarization in today's global scene can be daunting with its flood of multilingual content and varied viewpoints from different sources.
However, current studies often neglect such real-world scenarios as they tend to focus solely on either single-language or single-document tasks.
To bridge this gap, we aim to unify \textbf{M}ulti-lingual, \textbf{C}ross-lingual and \textbf{M}ulti-document \textbf{S}ummarization into a novel task, i.e., \textbf{\taskname}, which encapsulates the real-world requirements all-in-one.
Nevertheless, the lack of a benchmark inhibits researchers from adequately studying this invaluable problem.
To tackle this, we have meticulously constructed the {\datasetname} dataset by first collecting a wealth of multilingual news reports and restructuring them into event-centric format.
Additionally, we introduce the method of protocol-guided prompting for high-quality and cost-effective silver summary annotation.
In {\taskname}, we also highlight the challenge of \textit{conflicts} between news reports, in addition to the issues of \textit{redundancies} and \textit{omissions}, further enhancing the complexity of {\datasetname}.
Through extensive experimental analysis, we validate the quality of our dataset and elucidate the inherent challenges of the task.
We firmly believe that {\datasetname}, given its challenging nature, will greatly contribute to the multilingual communities and the evaluation of LLMs\footnote{Our dataset and code can be found at: \url{https://github.com/YYF-Tommy/GlobeSumm}.}. 
\end{abstract}
\section{Introduction}\label{sec:intro}
Summarization is a long-standing task in natural language processing (NLP) research~\citep{old-summ}. 
In recent years, significant advancements have been made in the field thanks to the rapid development of large language models (LLMs)~\citep{zhao2023survey, liu2023pre, dong2022survey, wei2022emergent, wei2022chain, shanahan2022talking}.
While LLMs have effectively addressed many traditional text summarization tasks~\citep{adams2023sparse, goyal2022news, pu2023summarization, zhang2023benchmarking}, the rapid globalization of information dissemination has created new demands for summarization techniques that can effectively summarize a large collection of event-centric multilingual news articles worldwide.

Events involved with armed conflicts, international relations, and political elections have always fascinated people worldwide. 
However, relying solely on news articles in a single language to gain an in-depth understanding of such events can be limiting.
This is because news reports from different countries are often influenced by their national standpoints and cultural biases, resulting in potential distortions~\citep{boykoff2004balance, baum2009war, baumeister2013distortions}. 
To obtain a more comprehensive insight into these events, it is crucial to explore news articles from various countries and languages, allowing us to consider diverse perspectives and access more objective information. 
Surprisingly, while advancements in LLMs have shown promising results in many NLP tasks, little research has been conducted for such real-world scenarios. 

To this end, we present the task of {\taskname} that unifies \textbf{M}ulti-lingual, \textbf{C}ross-lingual and \textbf{M}ulti-document \textbf{S}ummarization into a more general setting, aiming to align better with the multifaceted requirements in real-world scenarios.
The goal of {\taskname} is to succinctly capture the key information from a collection of documents written in various languages and present a cohesive summary in the target language.
Notably, the {\taskname} task has three distinctive features: (1) the input consists of multiple documents, (2) the multiple documents are in different languages, and (3) the multiple documents revolve around the same event.
However, the absence of a dataset that encompasses such features inhibits researchers from further study.

\begin{figure*}[t]
    \centering
    \includegraphics[width=1\textwidth]{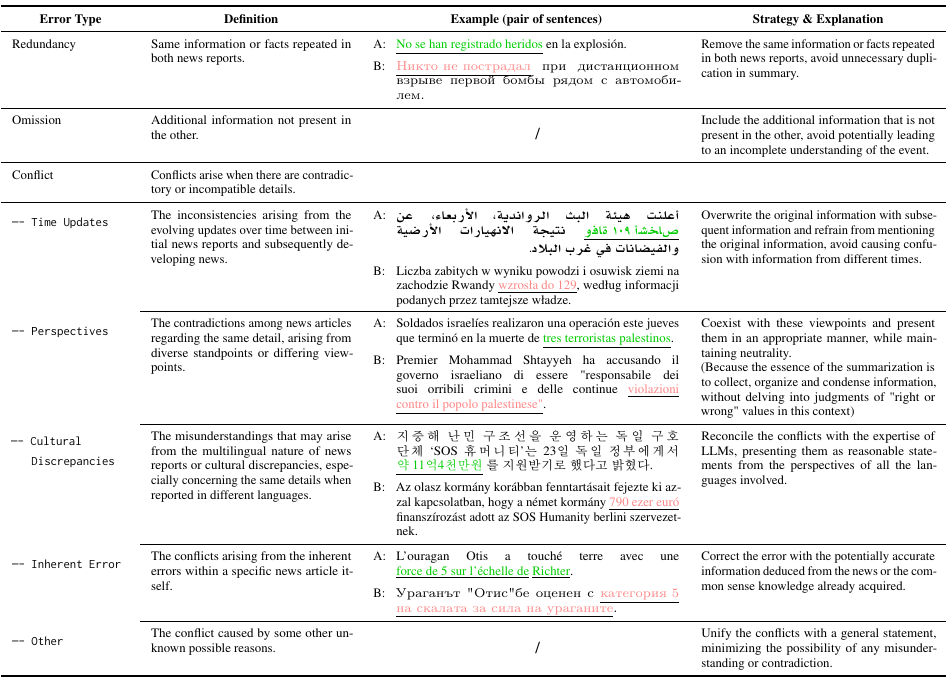}
    \caption{The protocol, we formulated for MCMS task, includes definitions of all potential error types, along with corresponding real-life examples for each type and approximate resolution strategies. (\textbf{English version in Figure~\ref{fig:error_types_en}})}
    \label{fig:error_types}
    \vspace{-0.3\baselineskip}
\end{figure*}

To close this gap, we meticulously construct the {\datasetname} dataset, which comprises the following two parts.
\textbf{Regarding news collection}, we begin by collecting a massive amount of news data from the GDELT database\footnote{\url{https://www.gdeltproject.org/}}, followed by a careful event-centric reranking and filtering process.
\textbf{Regarding silver-quality summary annotation}, we introduce protocol-guided prompting for high-quality and cost-effective silver-quality summary annotation. 
Specifically, based on extensive manual observations, we first develop a protocol, which takes into account three main challenges of {\taskname}: \textit{redundancies}, \textit{omissions}, and \textit{conflicts} (Figure~\ref{fig:error_types}), providing their definitions, examples, resolution strategies, and other relevant information.
The protocol-guided prompting method then requires LLMs to follow the established guidelines in the protocol during summary generation, which demonstrates performance close to or even surpassing human annotators (high-quality) and reduces the burden of manual annotation (cost-effective).

Building upon these foundations, our extensive experiments and analysis serve to validate the high quality of {\datasetname} and, more importantly, highlight the inherent challenges of dealing with \textit{redundancies}, \textit{omissions}, and \textit{conflicts} within {\taskname} task.
Notably, our discoveries emphasize that addressing conflicts arising from diverse perspectives significantly contribute to navigating information from different sources for high-performing LLMs.

\begin{figure*}[t]
    \centering
    \includegraphics[width=1\textwidth]{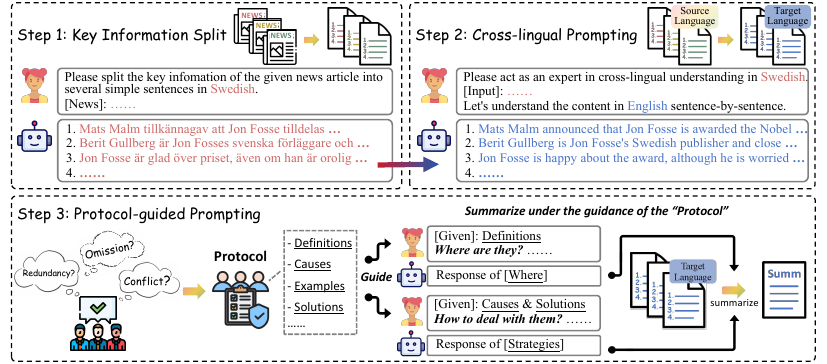}
    \caption{Overview of our silver-quality summary annotation methodology. The method consists of key information split, cross-lingual prompting and protocol-guided prompting.}
    \label{fig:pipeline}
\end{figure*}

\section{The {\datasetname} Dataset}
\subsection{Data Collection}
We gather news reports from countries and regions across 26 languages worldwide. 
This is accomplished by utilizing the news URLs provided in Google GDELT 2.0 project\footnote{\url{https://blog.gdeltproject.org/gdelt-2-0-our-global-world-in-realtime/}}.
We exclusively utilize news data from May 2023 to October 2023 for our benchmark to avoid introducing prior knowledge from web-crawled articles used in pre-training large language models\footnote{The cutoff of \texttt{GPT-4-1106-preview} is April 2023.}.

\subsection{Source Data Construction} \label{subsec:source}
One setting in {\taskname} is noteworthy that the multiple news reports within the same round input should be highly relevant to the same news event, rather than an open-domain task.
To address this, we employ a method involving event retrieval and manual verification to restructure the news reports.

\paragraph{Event Retrieval} 
To pinpoint news related to specific events, we leverage Wikipedia's current events portal\footnote{\url{https://en.wikipedia.org/wiki/Portal:Current\_events}} as a seed set. Each event in this set serves as a query input for our retrieval process. Our goal is to identify highly relevant news reports from the multilingual corpus. Initially, we translate the query event (originally in English) into multiple languages\footnote{We translate the queries by Google translation API}. Subsequently, we employ the BM25 retriever in \citet{Lin_etal_SIGIR2021_Pyserini} for retrieval in the respective language corpora, searching for the most query-relevant news articles.

\paragraph{Manual Verification} 
The retrieved news articles in different languages are supposed to be highly relevant to the provided description, but \textit{high relevance does not necessarily imply that they all present the same news event}.
Hence, we incorporate a post-retrieval manual verification process (see Appendix~\ref{appendix:A-1}). 

\subsection{Silver Summary Annotation Methodology}\label{subsec:method}
Next, we will elaborate on how we craft our silver-quality summaries in {\datasetname} (all the prompts can be found in Appendix~\ref{appendix:A-2}). 

\paragraph{Chronological Recurrent Summarization (CRS)} 
Our summary annotation approach is conducted under the CRS schema, aiming to distill key information from news articles in chronological order. 

Specifically, we begin by organizing these news documents in order of their respective timestamps. Then the summarization process is initiated by generating a concise summary for the first two articles. The obtained summary is then integrated with the subsequent article, and iteratively throughout the whole document set. 
CRS delivers a concise, timely summary by capturing the dynamic narrative and providing a comprehensive overview of the evolving information landscape in news articles.

\paragraph{Step 1: Key Information Split (KIS)} 
The large input length of a whole document set, averaging nearly 12K tokens in \datasetname, poses a great obstacle in {\taskname}. Therefore, we introduce the method of KIS to reduce the length of input by organizing key information from each document into several finely-grained sentences before summarizing the whole document set. 

\paragraph{Step 2: Cross-lingual Prompting (CLP)} 
Achieving cross-lingual alignment poses another fundamental challenge in multi- and cross-lingual tasks. To effectively capture the alignment from various input languages to target language, we employ cross-lingual alignment prompting method, which was first introduced in \citet{qin2023cross}. 

\begin{table*}[t]
    \centering
    \vspace{5pt}
    \scriptsize
    \begin{adjustbox}{width=\textwidth}
        \begin{tabular}{ccccccccc}
            \toprule
            \textbf{Dataset}  & \textbf{Domain} & \textbf{Multi-lingual} & \textbf{Cross-lingual} & \textbf{Multi-document} & \textbf{Focus} & \textbf{\# Document} & \textbf{\# Summary} & \textbf{\# Language} \\
            \hline
            MeetingBank~\citep{hu-etal-2023-meetingbank} & Meeting & {\redcross} & {\redcross} & {\redcross} & Redundancy & 1366 & 1366 & 1 \\ 
            MSAMSum~\citep{feng-etal-2022-msamsum} & Dialogue & {\greentick} & {\redcross} & {\redcross} & {/} & 5929 & 5929 & 6 \\
            MLSUM~\citep{scialom-etal-2020-mlsum} & News & {\greentick} & {\redcross} & {\redcross} & {/} & 1.5 millions & 1.5 millions & 5\\
            XL-Sum~\citep{hasan-etal-2021-xl} & News & {\redcross} & {\greentick} & {\redcross} & {/} & 1 million & 1 million & 44 \\
            WikiLingua~\citep{ladhak-etal-2020-wikilingua} & Wiki & {\greentick} & {\greentick} & {\redcross} & {/} & 140000 + & 770000 + & 18 \\
            Multi-News~\citep{fabbri-etal-2019-multi} & Wiki & {\redcross} & {\redcross} & {\greentick} & {/} & 250000 + & 50000 + & 1 \\
            OPENASP~\citep{amar2023openasp} & News & {\redcross} & {\redcross} & {\greentick} & {Open aspect} & 13582 & 1,361 & 1 \\
            \noalign{\vskip 0.5ex}\cdashline{1-9}\noalign{\vskip 0.5ex}
            \datasetname & News & {\greentick} & {\greentick} & {\greentick} & \makecell[c]{Redundancy, Omission, Conflict} & 4687 & 4317 & 26 \\
            \hline
        \end{tabular}
    \end{adjustbox}
    \caption{Comparisons with existing Multi-lingual, Cross-lingual or Multi-document summarization datasets.}
    \label{tab:dataset_cmp}
\end{table*}

\paragraph{Step 3: Protocol-guided Prompting (PGP)} 
We first introduce the method of \textit{protocol-guided prompting} (PGP) to achieve high-quality summary annotation.
Based on our manual observation of diverse news articles across multiple languages and documents, we have concluded three primary hurdles in {\taskname}: \textit{redundancies}, \textit{omissions}, and \textit{conflicts}. 
The details shown in Figure~\ref{fig:error_types}, which constitute our protocol, will be incorporated as part of the prompt to assist the LLMs in more effectively identifying and handling these hurdles while summarizing the documents. 

More specifically, the procedure of how we address redundancies, omissions and conflicts can be broadly divided into two parts: (1) \textit{where are they?} and (2) \textit{how to deal with them?}

\textbf{(1) Where are they?} (\underline{\texttt{[where]}}) We furnish LLM with the definitions of redundancy, omission, and conflict (in Figure~\ref{fig:error_types}) and request LLM to adeptly pinpoint the occurrences of these issues between documents based on provided definitions. 

\textbf{(2) How to deal with them?} (\underline{\texttt{[strategies]}})
As shown in our protocol (refer to Figure~\ref{fig:error_types}), we have conducted a manual synthesis and conclusion for these issues, especially conflicts. Based on the various causes that may give rise to these problems, we have elegantly formulated different 
solutions. Then, we request LLM to delineate specific strategies for each conflict arising from different reasons in the actual scenarios, following the customized general solutions we have outlined.

With the assistance of the knowledge in our protocol, we effectively achieve the two subtasks of \underline{\texttt{[where]}} and \underline{\texttt{[strategies]}}. 
Consequently, LLM's responses to \underline{\texttt{[where]}} and \underline{\texttt{[strategies]}} are utilized to generate our silver summaries. The detailed implementation can be found in Table~\ref{tab:full_prompts} in the appendix.

\begin{table}[t!]
\centering 
\setlength\tabcolsep{3pt}

\resizebox{\columnwidth}{!}{%

\begin{tabular}{lrrr}

\toprule
\textbf{Dataset}  & \textbf{\# Event} & \textbf{\# Document} & \textbf{\# Summary} \\
\midrule
Total  &  &         \\
\noalign{\vskip 0.2ex}\cdashline{1-4}\noalign{\vskip 0.2ex}
Num & 370 & 4687 & 4317 \\
Avg Token Length & 11568.46 & 913.23 & 368.04 \\
\midrule 
Train Set Size  & 222 & 2848 & 2626\\
Valid Set Size  & 74 & 897 & 823\\
Test Set Size  & 74 & 942 & 868\\
\bottomrule
\end{tabular}
}
\caption{Statistics of the {\datasetname} dataset. The token length was calculated by \textit{tiktoken}\protect\footnotemark.}
\label{tab:statistics}
\end{table}
\footnotetext{\url{https://github.com/openai/tiktoken}}
\begin{table*}
	\centering
	\begin{adjustbox}{width=\textwidth}
		\begin{tabular}{l|cccccccccccc|c}
			\hline
			Method (\texttt{GPT-3.5-turbo}) & AR & DE & EL & EN & ES & HI & RO & RU & TH & TR & VI & ZH & AVG
			\\
			\hline
			\underline{KIS vs. Summarize} &&&&&&&&&&&& \\
			
			\texttt{Summarize}   & 34.7 & 49.4 & 37.7 & 53.5 & 49.1 & 34.4 & 49.2 & 33.7 & 38.7 & 42.0 & 43.1 & 51.4 & 43.1 \\
            \texttt{Summarize-Extend}   & 45.6 & 64.5 & 51.0 & 65.0 & 64.1 & 44.7 & 57.5 & 45.2 & 44.7 & 50.7 & 57.5 & 63.9 & 54.5 \\
			KIS &  \textbf{50.9} & \textbf{66.9} & \textbf{55.0} & \textbf{74.5} & \textbf{69.4} & \textbf{50.9} & \textbf{65.8} & \textbf{50.9} & \textbf{50.5} & \textbf{51.9} & \textbf{62.5} & \textbf{67.7} & \textbf{59.8}\\
			
			\hline
			\underline{CLP vs. Translate} &&&&&&&&&&&& \\
			\texttt{Translate-En} & 51.3 & 62.9 & 60.4 & \scalebox{1}- & 62.6 & \textbf{60.2} & 62.9 & 54.7 & \textbf{48.7} & 60.2 & 58.7 & 49.5 & 57.5 \\
			CLP-En & \textbf{55.0} & \textbf{67.1} & \textbf{60.6} & \scalebox{1}- & \textbf{66.8} & 56.6 & \textbf{64.8} & \textbf{59.0} & 47.0 & \textbf{61.9} & \textbf{60.2} & \textbf{55.6} & \textbf{59.5} \\
			\hline
		\end{tabular}
	\end{adjustbox}
	\caption{
		The Acc. performance of \underline{KIS vs. Summarize} and \underline{CLP vs. Translate} on XQuAD.
	}
	\label{tab:versus}
\end{table*}

\subsection{Statistics}
Following the methodology described in Section~\ref{subsec:method}, our silver-quality summaries are generated with GPT-4 model\footnote{All GPT-4 mentioned in this paper refer to GPT-4-1106-preview. In order to significantly reduce costs, only the PGP phase is handled by GPT-4, while KIS and CLP process are executed by GPT-3.5-turbo-16k.} as the backbone. 

A total of 370 news events, consisting of 4687 news articles, have been finally retained in {\datasetname}. 
The entire dataset spans 26 languages and each news event is associated with a minimum of 10 news reports in different languages, adding to the challenge of our dataset. 
Due to the recurrent nature of CRS schema (Section~\ref{subsec:method}), {\datasetname} offers silver summaries for document subsets of any size within the whole collection of documents related to the same event, totaling 4317 in number.
And GPT-4's responses to \underline{\texttt{[where]}} and \underline{\texttt{[strategies]}} are also available in {\datasetname}.
The language distribution can be found in Table~\ref{tab:lang_dis}. 

As shown in Table~\ref{tab:dataset_cmp}, {\datasetname} stands out for being multi-lingual, cross-lingual, and multi-document and focuses on addressing redundancies, omissions, and conflicts. These qualities make {\datasetname} distinctive and practically valuable.

We split {\datasetname} into train, validation and test sets (Table~\ref{tab:statistics}). 
Subsequent experiments (Section~\ref{sec:exp}) are carried out on the test set.
Our expenses can be found in Appendix~\ref{appendix:D}.
\section{Annotation Quality Assessment}
We next examine the superiority of our annotation method.
\begin{table}[t!]
    \centering 
    \resizebox{\columnwidth}{!}{%
        \begin{tabular}{c|ccc|cc}
            \toprule
            \multicolumn{6}{c}{\textit{Inter-annotator Agreement Scores}} \\
            \hline
            \textbf{$Issue$} & \textbf{$Annotator_1$} & \textbf{$Annotator_2$} & \textbf{$Annotator_3$} & \textit{\textbf{Kappa}} & \textit{\textbf{Agreement}} \\
            \hline
            $Redundancy$ & 197 & 194 & 189 & \underline{0.93} & \underline{0.96} \\ 
            $Omission$ & 491 & 482 & 481 & \underline{0.95} & \underline{0.98} \\ 
            $Conflict$ & 45 & 44 & 39 & \underline{0.86} & \underline{0.93} \\
            \hline
        \end{tabular}}
    \caption{The number of identified issues by annotators, along with their inter-annotator agreement scores.}
\label{tab:agreement}
\end{table}
\begin{figure}[t]
    \centering
    \includegraphics[width=1\linewidth]{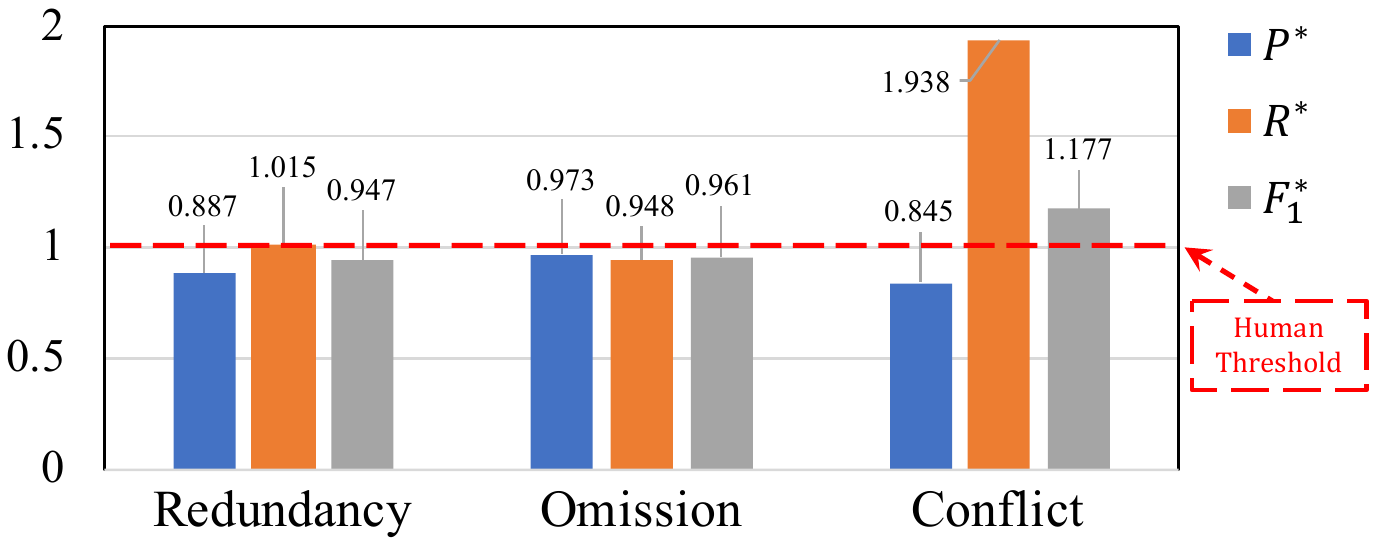}
    \caption{$P^*$, $R^*$ and $F_1^*$ scores of GPT-4 in \underline{\texttt{[where]}}. All values are calculated in micro-averaging.}
    \label{fig:prf&agreement}
\end{figure}

\subsection{Compete with Human Annotation} \label{subsec:GPT4}
In this section, we evaluate how well GPT-4 addresses \underline{\texttt{[where]}} and \underline{\texttt{[strategies]}} under the guidance of our protocol by comparing its performance with human annotation.

We randomly select 50 pairs of documents in \datasetname, with each pair focusing on the same event. Next, we invite 3 human annotators to identify the redundancies, omissions, and conflicts between pairs of documents. The high inter-annotator agreement in Table~\ref{tab:agreement} exhibits the reliability of our annotated data, which will serve as the standard for evaluating the performance of GPT-4.

$P^*$, $R^*$ and $F_1^*$ (see detailed formulas in Appendix~\ref{appendix:B-1}), the variants of Precision, Recall, and F\textsubscript{1} metrics, are utilized for evaluation\footnote{The $R^*$ value here may exceed 1, as GPT-4 has demonstrated the ability to identify additional redundancies, omissions, and conflicts overlooked by human annotators. However, through manual verification, some of these overlooked items are also confirmed as accurate answers, thus contributing to the $R^*$ metric numerator during calculation.}.

The scoring results in Figure~\ref{fig:prf&agreement} shows that GPT-4 performs comparably to human annotators in terms of identifying \textbf{redundancy} and \textbf{omission}, with $F_1^*$ scores approaching human threshold (value 1). 
Regarding \textbf{conflict}, GPT-4 outperforms human annotators in $F_1^*$ scores, and its $R^*$ value achieves nearly double that of human annotators. 
The results strongly indicate that guided by our protocol, GPT-4 can effectively replace or even surpass humans in completing the subtask \texttt{[\underline{where}]}.

In subtask \texttt{[\underline{strategies}]}, compared to addressing redundancies and omissions, resolving conflicts is evidently more complex and challenging. Therefore, in this study, we conduct a manual evaluation of the 93 conflict resolution strategies generated by GPT-4 for those 50 pairs of samples. The outcome reveals a 96.8\% accuracy (90 out of 93), indicating that GPT-4 consistently generates correct, reasonable, and protocol-compliant strategies.

\subsection{Component-wise Analysis} \label{subsec:component}
Next, we will explore where the advantages of KIS and CLP are specifically manifested.
We conduct comparative experiments on XQuAD \citep{Artetxe:etal:2019, dumitrescu2021liro}, exploring KIS versus Summarize and CLP versus Direct Translate (see detailed implementation in Appendix~\ref{appendix:B}).

\paragraph{(1) KIS results better condensing quality.}
As shown in Table~\ref{tab:versus}, we find that KIS exhibits a remarkable superiority over \texttt{Summarize} across all languages (with 16.7\% improvements on average accuracy), strongly indicating that the context after KIS is more comprehensible for LLMs compared to the summarized context. 
Recognizing the impact of compression ratios on the total information provided (KIS\textasciitilde 343 tokens; \texttt{Summarize}\textasciitilde 242 tokens), we introduce another control group named \texttt{Summarize-Extend} with a longer compressed context (averaging 579 tokens). 
Nevertheless, KIS still outperforms \texttt{Summarize-Extend} by 5.3\% in accuracy, further illustrating that KIS is a better method for capturing the key information.

\paragraph{(2) CLP outperforms direct translation in cross-lingual alignment.}
As depicted in Table~\ref{tab:versus}, CLP demonstrates higher accuracy than \texttt{Translate} by averaging 2.0\%, which illustrates that CLP can assist LLMs more effectively in achieving semantic alignment between languages, thereby enhancing cross-lingual comprehension. As verified in \citet{qin2023cross}, CLP is not a vanilla translation but utilizes the cross-lingual semantic alignment.

\section{Experiments}\label{sec:exp}

\subsection{Baselines}
Our experiments utilize various baselines, each composed of a combination of "schema + pipeline".

\noindent\paragraph{Schemas} 
To validate the effectiveness of Chronological Recurrent Summarization (CRS), we investigate the two schemas for comparison: 
(a) \textit{Single-turn Summarization} summarizes a document set within a single-turn generation; 
(b) \textit{Chronological Recurrent Summarization} iteratively summarizes two documents at a time in a time-ordered manner.

\noindent\paragraph{Pipelines} To further validate the advantages of KIS and CLP in addressing lengthy inputs and cross-language understanding, we conduct comparative tests with these commonly used methods: 
(a) \textit{Translate-then-Summarize}; 
(b) \textit{Summarize-then-Translate}; 
(c) \textit{KIS-then-CLP}.

Similarly, we conduct experiments using two different approaches for summarization:
(a) \textit{Direct Summarization};
(b) \textit{Protocol-guided Prompting}.

Detailed introductions to these pipelines can be found in Appendix~\ref{appendix:C-1}

\paragraph{Models} 
We select three representative LLMs that feature long context capability, each of which supports at least a 16k context window.
\begin{itemize}[leftmargin=*,topsep=0pt]
\setlength{\parsep}{0pt}
\setlength{\parskip}{0pt}
\item \textbf{GPT-3.5-turbo-16k} is an advanced GPT-3.5 series model with a 16k context window. 
\item \textbf{Vicuna-7B-v1.5-16k} \citep{zheng2023judging} is an open-source language model fine-tuned from Llama2, and supports a 16k context window.
\item \textbf{ChatGLM3-6B-32k} \citep{du2022glm} is an open-source language model based on General Language Model (GLM) framework, and supports a 32K context window.
\end{itemize}

\subsection{Metrics}
We evaluate the quality of the generated summaries using following metrics (see Appendix~\ref{appendix:C-2} for detailed definitions and formulas):
\begin{itemize}[leftmargin=*,topsep=0pt]
\setlength{\parsep}{0pt}
\setlength{\parskip}{0pt}
\item \textbf{\textsc{ROUGE}} \citep{lin2004rouge} measures the overlap co-occurrence of n-grams between the candidate and reference summaries.
\item \textbf{Red} \citep{chen-etal-2021-training} is a self-referenced metric for \textit{redundancy} evaluation.
\item \textbf{Normalized Inverse of Coverage (NIC)} captures \textit{Omission}, as the inverse of a coverage of key information from reference summary.
\item \textbf{Conflict Resolution Effectiveness (CRE)} metric evaluates how well a candidate summary addresses \textit{conflict}.
\end{itemize}

\begin{table*}
	\centering
	\begin{adjustbox}{width=\textwidth}
        \begin{tabular}{l|cccc|cccc|cccc}
            \hline
            \multirow{2}{*}{Schema \& Pipeline} & \multicolumn{4}{c|}{GPT-3.5-turbo-16k}  & \multicolumn{4}{c|}{Vicuna-7b-v1.5-16k} & \multicolumn{4}{c}{Chatglm3-6b-32k} \\
            \cline{2-13}
            & R-L {\uparr} & Red {\downarr} & NIC {\downarr} & CRE {\uparr} & R-L {\uparr} & Red {\downarr} & NIC {\downarr} & CRE {\uparr} & R-L {\uparr} & Red {\downarr} & NIC {\downarr} & CRE {\uparr} \\
            \noalign{\vskip 0.2ex} \hline \noalign{\vskip 0.2ex}
            \underline{Single-turn Summaization (STS)} &&&&&&&&&  \\
            \texttt{Translate-then-Summarize + Direct}      & 19.86	& \textbf{30.06}	& 84.77	& 56.33	& 18.49	& \textbf{29.38}	& 87.12	& 55.20	& 18.98	& 32.49	& 85.17	& \textbf{58.43} \\
            \texttt{Summarize-then-Translate + Direct}      & 20.07	& 30.57	& 84.59	& 56.18	& 19.08	& 30.70	& 87.44	& 54.62	& 19.45	& 32.82	& 86.87	& 56.05 \\
            \texttt{KIS-then-CLP + Direct}                  & 21.25	& 30.91	& 82.05	& 59.02	& 18.96	& 30.73	& 87.19	& 53.37	& 19.27	& 35.06	& 85.08	& 57.11 \\
            \noalign{\vskip 0.2ex}\cdashline{1-13}\noalign{\vskip 0.2ex}
            \texttt{Translate-then-Summarize + Protocol}    & 19.41	& 31.20	& 89.34	& 54.29	& 19.07	& 30.09	& 87.82	& 53.06	& 18.76	& 30.33	& 87.20	& 55.51 \\
            \texttt{Summarize-then-Translate + Protocol}    & 19.18	& 31.04	& 87.79	& 54.23	& 18.69	& 29.43	& 86.73	& 55.22	& 18.76	& 31.44	& 88.10	& 53.76 \\
            \texttt{KIS-then-CLP + Protocol}                & 21.17	& 31.63	& 80.01	& 54.01	& 18.82	& 33.10	& 88.24	& 57.00	& 19.29	& \textbf{30.15}	& 89.19	& 57.20 \\
            
            \noalign{\vskip 0.2ex} \hline \noalign{\vskip 0.2ex}
            \underline{Chronological Recurrent Summarization (CRS)} &&&&&&&&& \\
            \texttt{Translate-then-Summarize + Direct}      & 20.27	& 33.36	& 82.06	& 54.58	& 20.14	& 32.37	& 77.99	& 56.28	& 20.08	& 32.80	& \textbf{77.94}	& 55.95 \\
            \texttt{Summarize-then-Translate + Direct}      & 20.15	& 32.86	& 80.81	& 55.47	& 19.62	& 33.49	& 82.73	& 57.23	& 20.13	& 33.39	& 84.21	& 53.62 \\
            \texttt{KIS-then-CLP + Direct}                  & 21.92	& 34.29	& 74.35	& 56.45	& 20.59	& 33.60	& 80.38	& 53.39	& 20.12	& 34.65	& 82.11	& 55.11 \\
            \noalign{\vskip 0.2ex}\cdashline{1-13}\noalign{\vskip 0.2ex}
            \texttt{Translate-then-Summarize + Protocol}    & 21.14	& 30.99	& 76.21	& 55.71	& 20.85	& 32.67	& 80.30	& 54.52	& \textbf{20.48}	& 31.69	& 79.06	& 55.38 \\
            \texttt{Summarize-then-Translate + Protocol}    & 21.24	& 31.32	& 81.08	& 54.55	& 20.21	& 31.67	& 80.80	& 57.51	& 19.60	& 32.88	& 85.10	& 54.49 \\
            \texttt{KIS-then-CLP + Protocol}                & \textbf{22.06} & 32.30	& \textbf{70.09}	& \textbf{59.11}	& \textbf{20.94}	& 34.92	& \textbf{76.62}	& \textbf{58.24}	& 20.19	& 33.28	& 85.15	& 54.86 \\
            
            \noalign{\vskip 0.2ex} \hline \noalign{\vskip 0.2ex}
            \noalign{\vskip 0.2ex} \hline \noalign{\vskip 0.2ex}
            \texttt{Translate-then-Summarize.Avg}           & 20.17	& \textbf{31.40}	& 83.09	& 55.23	& 19.64	& \textbf{31.13}	& 83.31	& 54.77	& 19.58	& \textbf{31.83}	& \textbf{82.34}	& \textbf{56.32} \\
            \texttt{Summarize-then-Translate.Avg}           & 20.16	& 31.45	& 83.57	& 55.11	& 19.40	& 31.32	& 84.42	& \textbf{56.15}	& 19.49	& 32.63	& 86.07	& 54.48 \\
            \texttt{KIS-then-CLP.Avg}                       & \textbf{21.60	}& 32.28	& \textbf{76.62}	& \textbf{57.15}	& \textbf{19.83}	& 33.09	& \textbf{83.11}	& 55.50	& \textbf{19.72}	& 33.29	& 85.38	& 56.07 \\
            \noalign{\vskip 0.2ex} \hline \noalign{\vskip 0.2ex}
            \texttt{STS.Avg}                                & 20.16	& \textbf{30.90}	& 84.76	& 55.68	& 18.85	& \textbf{30.57}	& 87.42	& 54.75	& 19.09	& \textbf{32.05}	& 86.94	& \textbf{56.34} \\
            \texttt{CRS.Avg}                                & \textbf{21.13}	& 32.52	& \textbf{77.43}	& \textbf{55.98}	& \textbf{20.39}	& 33.12	& \textbf{79.80} & \textbf{56.20}	& \textbf{20.10}	& 33.12	& \textbf{82.26}	& 54.90 \\
            \noalign{\vskip 0.2ex} \hline \noalign{\vskip 0.2ex}
            \texttt{STS + Direct.Avg}                       & \textbf{20.39}	& \textbf{30.51}	& \textbf{83.80}	& \textbf{57.18}	& 18.84	& \textbf{30.27}	& \textbf{87.25}	& 54.40	& \textbf{19.23}	& 33.46	& \textbf{85.71}	& \textbf{57.20} \\
            \texttt{STS + Protocol.Avg}                     & 19.92	& 31.29	& 85.71	& 54.18	& \textbf{18.86}	& 30.87	& 87.60	& \textbf{55.09}	& 18.94	& \textbf{30.64}	& 88.16	& 55.49 \\
            \noalign{\vskip 0.2ex} \hline \noalign{\vskip 0.2ex}
            \texttt{CRS + Direct.Avg}                       & 20.78	& 33.50	& 79.07	& 55.50	& 20.12	& 33.15	& 80.37	& 55.63	& \textbf{20.11}	& 33.61	& \textbf{81.42}	& 54.89 \\
            \texttt{CRS + Protocol.Avg}                     & \textbf{21.48}	& \textbf{31.54}	& \textbf{75.79}	& \textbf{56.46}	& \textbf{20.67}	& \textbf{33.09}	& \textbf{79.24}	& \textbf{56.76}	& 20.09	& \textbf{32.62}	& 83.10	& \textbf{54.91} \\
            \hline
        \end{tabular}
    \end{adjustbox}
    \caption{
		Evaluation results for all configurations of schemas and pipelines on different LLMs. "\texttt{Direct}" indicates direct summarization, while "\texttt{Protocol}" represents summarization with protocol-guided prompting. {\uparr} denotes higher score the better and {\downarr} means the opposite. \texttt{X.Avg} represent the average performance of all \texttt{X-based} baseline.
	}
	\label{tab:main}
\end{table*}

\subsection{Main Results}\label{subsec:main_results}

The main results are illustrated in Table~\ref{tab:main} (see Table~\ref{tab:rouge_main} in Appendix for full ROUGE results). From the results, we have the following observations:

\paragraph{(1) Omissions and Conflicts mitigated, yet Redundancies persist.}
As shown in Table~\ref{tab:main}, unlike omissions and conflicts, which can be mitigated with the introduction of our methodology (CRS, KIS, CLP and PGP), redundancies, on the contrary, tends to persist, even exacerbate. The results across all three models do not seem to reflect the effectiveness of our approach in addressing redundancy. This divergence on different issues emphasizes the multifaceted nature of {\taskname}.

\paragraph{(2) Preferential performance in CRS with Protocol-guided Prompting.} 
From the results on GPT-3.5-turbo-16k and Vicuna-7b-v1.5-16k as illustrated in Table~\ref{tab:main}, we find that protocol-guided prompting outperforms \texttt{Direct} only under the CRS schema, while its superiority is not evident under STS. This is within our expectations, as STS requires LLMs to simultaneously identify and coordinate redundancies, omissions, and conflicts across all news documents, while CRS simplifies summarization by focusing on two documents at a time. 

\paragraph{(3) LLM's Sensitivity to Protocol-guided Prompting.}
Protocol-guided prompting demonstrates certain advantages on both GPT-3.5-turbo-16k and Vicuna-7b-v1.5-16k in Table~\ref{tab:main}. However, with Chatglm3-6b-32k model, regardless of STS or CRS schema, protocol-guided prompting underperforms direct summarization. This indicates that the effectiveness of protocol-guided prompting depends on the model's capabilities, which requires understanding relatively complex prompts.

\section{Further Analysis}\label{sec:analysis}

\begin{table}[t]
    \resizebox{\columnwidth}{!}{%
        \begin{tabular}{l|cccc}
            \hline
            \multirow{2}{*}{Schema \& Pipeline} & \multicolumn{4}{c}{GPT-3.5-turbo-16k} \\
            \cline{2-5}
            & R-L {\uparr} & Red {\downarr} & NIC {\downarr} & Con {\uparr} \\
            \noalign{\vskip 0.2ex} \hline \noalign{\vskip 0.2ex}
            \underline{Single-turn Summaization (STS)} &&&&  \\
            \texttt{None + Direct}            & 17.60 & \textbf{30.31} & 85.01 & 49.52 \\
            \texttt{KIS-only + Direct}        & 19.60 & 30.82 & 86.26 & 55.58 \\
            \texttt{KIS-then-CLP + Direct}    & \textbf{21.25} & 30.91 & \textbf{82.05} & \textbf{59.02} \\
            \noalign{\vskip 0.2ex}\cdashline{1-5}\noalign{\vskip 0.2ex}
            \texttt{None + Protocol}          & 19.12 & 31.39 & 89.42 & 44.20 \\
            \texttt{KIS-only + Protocol}      & 19.00 & \textbf{30.91} & 87.87 & \textbf{55.47} \\
            \texttt{KIS-then-CLP + Protocol}  & \textbf{21.17} & 31.63 & \textbf{80.01} & 54.01 \\
            \noalign{\vskip 0.2ex} \hline \noalign{\vskip 0.2ex}
            \underline{Chronological Recurrent Summarization (CRS)} &&&& \\
            \texttt{None + Direct}            & 18.45 & \textbf{31.58} & 84.26 & 53.63 \\
            \texttt{KIS-only + Direct}        & 20.76 & 33.67 & 78.03 & 52.68 \\
            \texttt{KIS-then-CLP + Direct}    & \textbf{21.92} & 34.29 & \textbf{74.35} & \textbf{56.45} \\
            \noalign{\vskip 0.2ex}\cdashline{1-5}\noalign{\vskip 0.2ex}
            \texttt{None + Protocol}          & 19.47 & \textbf{29.46} & 82.72 & 53.94 \\
            \texttt{KIS-only + Protocol}      & 21.46 & 31.05 & 75.51 & 56.62 \\
            \texttt{KIS-then-CLP + Protocol}  & \textbf{22.06} & 32.30 & \textbf{70.09} & \textbf{59.11} \\
            \hline
        \end{tabular}}
        \caption{
    		The evaluation results of ablation studies on \textit{KIS-then-CLP} stage.
    	}
	\label{tab:ablation}
        \vspace{-0.5\baselineskip}
    
\end{table}

\subsection{Ablation Study}
We conduct ablation studies to investigate the effect of the \textit{KIS-then-CLP} stage, as shown in the rows of Table~\ref{tab:ablation} (see Table~\ref{tab:rouge_ablation} for full ROUGE results).

We observe in Table~\ref{tab:ablation} that GPT-3.5-turbo-16k exhibits a noticeable performance decline in summarization when either the KIS or CLP steps are omitted. This also indicates that relying solely on the ability of LLMs themselves to handle long-text and multi-lingual inputs may not be an appropriate solution at present, highlighting the necessity of pre-emptively explicit text compression and cross-language alignment for LLMs.

\begin{figure}[t]
    \centering
    \includegraphics[width=1\linewidth]{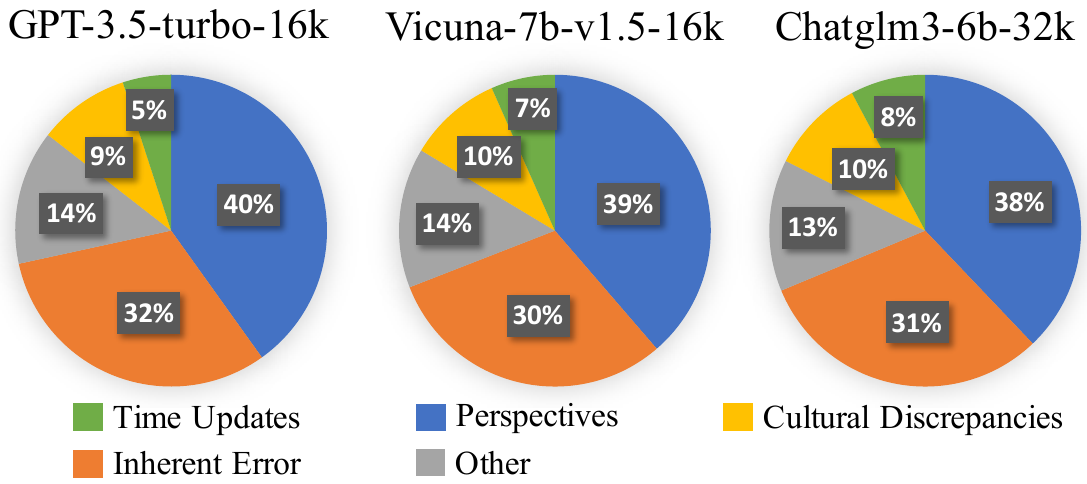}
    \caption{Average error rates of LLMs for each type of conflict as a proportion of the total errors.}
    \label{fig:pies}
    \vspace{-0.5\baselineskip}
\end{figure}

\subsection{Error Analysis}\label{subsec:types}
We further present the average error rates of LLMs for each type of conflict as a proportion of the total errors in Figure~\ref{fig:pies}.

The results illustrate that conflicts caused by diverse perspectives account for the majority of errors in LLMs' practical performance. This also reflects the ongoing challenge faced by current LLMs in efficiently processing and integrating information originating from a wide array of viewpoints and perspectives in complex real-world scenarios.

\subsection{LLM's Scale-Effect on PGP}\label{subsec:scale}
The sensitivity observation (Section \ref{subsec:main_results}) prompts our study into the llama2 \citep{touvron2023llama} series models with varying sizes (Appendix~\ref{appendix:C-3}).

We compare the performance of \textit{Direct Summarization} and \textit{Protocol-guided Prompting} (Table~\ref{tab:llama2}, \ref{tab:rouge_llama2}), the $\Delta$ results shown in Table~\ref{tab:llama2} exhibit favorable changes in both omission and conflict aspects as the model size increases (NIC {\downarr}: \(13.96 \rightarrow 5.01 \rightarrow 4.36\); CRE {\uparr}: \(- 4.96 \rightarrow 0.61 \rightarrow 3.32\)). This indicates that with the growth of model scale, protocol-guided prompting outperforms direct summarization, but redundancy remains an issue.

\begin{figure}[t]
    \centering
    \includegraphics[width=1\linewidth]{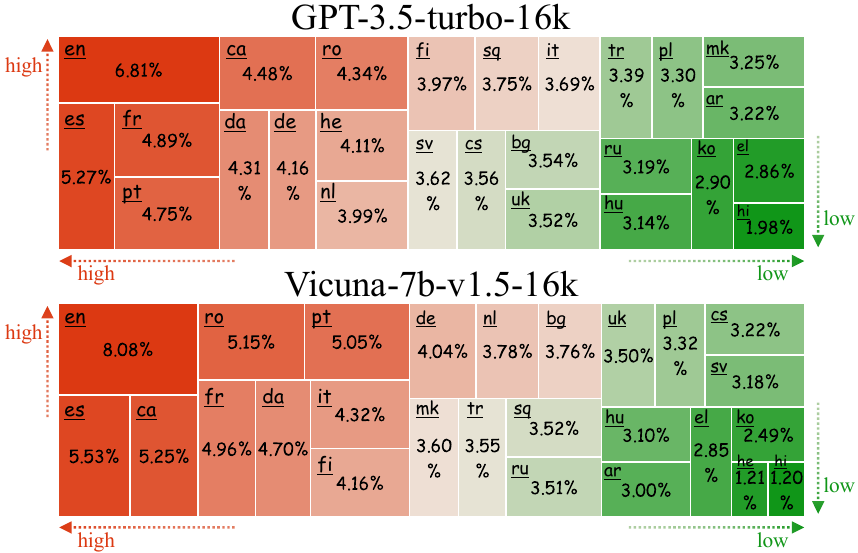}
    \caption{The average proportion of content in summaries generated by LLMs that is entailed in different source documents across 26 languages.}
    \label{fig:low_resource}
\end{figure}

\subsection{Apathy towards Low-Resource Languages}
Within {\taskname}, we undertake several experiments to investigate LLM's prejudices across various languages (details can be found in Appendix~\ref{appendix:C-3}).

The results on GPT-3.5-turbo-16k, Vicuna-7B-v1.5-16k in Figure~\ref{fig:low_resource} all indicate 
a tendency to prioritize content from documents in high-resource languages like English and Spanish, with only a small part from documents in low-resource languages, like Hindi, Greek, and Hebrew. This preference poses a challenge for current LLMs to be fair summarizers across all languages. 

\section{Related Work}
\subsection{Multi-lingual and Cross-lingual Summarization}
Multi-lingual summarization (MLS) aims to process documents in multiple languages and generate their summaries in the corresponding language. The MultiLing-2015 dataset \citep{giannakopoulos-etal-2015-multiling} initiates interest in this task, leading to increasing subsequent studies \citep{vanetik2015multilingual, litvak2016museec, cao2020multisumm}. Recently, with the availability of many large-scale MLS datasets \citep{varab-schluter-2021-massivesumm, hasan-etal-2021-xl, feng-etal-2022-msamsum}, notable progress is achieved one after another. 
Cross-lingual summarization (CLS) summarizes given documents in one language into summaries in another target language. 
The early work mainly focuses on pipeline methods \citep{yao2015phrase, ouyang2019robust, wan2010cross}, leading to error propagation. The recent large-scale CLS datasets \citep{zhu-etal-2019-ncls, wang-etal-2022-clidsum, zheng2023long} are shifting the research attention to end-to-end studies \citep{cao2020jointly, liang-etal-2022-variational}. 
Considering the close relation between MLS and CLS, \citet{feng-etal-2022-msamsum} evaluate the MLS models on CLS to show their zero-shot CLS ability, \citet{wang-etal-2023-towards-unifying} unifies MLS and CLS into a more general setting of many-to-many.
Unlike typical MLS and CLS tasks, {\taskname} involves multi-document summarization across multiple languages in a single input round, posing a greater challenge.

\subsection{Multi-document Summarization}
Multi-document summarization (MDS) refers to the task of summarizing the text in multiple documents into a concise summary. Previous studies have delved into various approaches, encompassing extractive \citep{angelidis-lapata-2018-summarizing, zheng-etal-2019-subtopic, mao-etal-2020-multi} and abstractive techniques \citep{gehrmann-etal-2018-bottom, lebanoff-etal-2018-adapting, zhang-etal-2018-adapting}. And researchers mainly focus on reducing the redundancy among documents \citep{peyrard-etal-2017-learning, xiao-carenini-2020-systematically, chen-etal-2021-training}. 
Currently, there is a growing focus on MDS tasks in more diverse settings.
\citet{zhou2023odsum} highlights the challenge of open-domain MDS, \citet{amar2023openasp} proposes aspect-based summarization in MDS to better fit the needs in real-world scenarios.
Our {\taskname} extends typical MDS task by incorporating a multi-lingual usage. Unlike prior MDS efforts that targeted redundancy reduction, {\taskname} also highlights the challenges of addressing omission and conflict between multipe documents, which is crucial for real-world information management across diverse sources.

\section{Conclusion}
To conclude, our study presents the task of {\taskname} that unifies \textbf{M}ulti-lingual, \textbf{C}ross-lingual and \textbf{M}ulti-document \textbf{S}ummarization to align better with the diverse needs in real-world scenarios.
Our benchmark, {\datasetname}, serves this demand as the first dataset for such scenario, offering high-quality summaries generated through protocol-guided prompting. Through experiments and analysis, conducted on outperforming LLMs, we unveil the shortcomings of LLMs in {\taskname} and highlight the challenges of addressing redundancies, omissions and conflicts. Overall, we believe {\datasetname} holds the potential to be used for evaluating the performance of LLMs in handling multi-lingual and multi-document tasks and the way we utilize protocol-guided prompting can serve as a practical case for cost-effective annotation.

\section*{Ethics Statement}
We utilize publicly available news data, which may contain viewpoints from different perspectives. The output results in the paper do not necessarily represent the views of the authors.
\section*{Limitations}
While our dataset is constructed with GPT-4, budget constraints prevent us from exploring further experimental results on the GPT-4 model. 

Our work primarily focuses on addressing redundancies, omissions, and conflicts among documents. However, in our attempts, we have found that while omissions and conflicts can be alleviated to some extent through our method, redundancies have not shown significant improvement.

Due to the recurrent nature of CRS, our reference summaries can cover any truncation length within the document set, as opposed to only providing a single final summary for each document set in many typical MDS datasets. However, in this work, there has not been an extensive investigation into this particular aspect, such as the impact of document quantity and language diversity on the difficulty of {\taskname}.

Gaining a profound understanding of a specific global news event involves more than the {\taskname} task discussed in our work. Exploring how to group news reports about the same event is also a worthwhile research endeavour. However, in the data construction phase of this study, the effectiveness of this step is ensured through manual post-validation without delving into its methodology.

\section*{Acknowledge}
Xiaocheng Feng is the corresponding author of this work. We thank the anonymous reviewers for their insightful comments. This work was supported by the National Natural Science Foundation of China (NSFC) (grant 62276078, U22B2059), the Key R\&D Program of Heilongjiang via grant 2022ZX01A32, the International Cooperation Project of PCL, PCL2022D01 and the Fundamental Research Funds for the Central Universities (Grant No.HIT.OCEF.2023018).

\bibliography{custom}

\appendix
\section{{\datasetname} Construction Details}\label{appendix:A}

\begin{figure*}[t]
    \centering
    \includegraphics[width=1\textwidth]{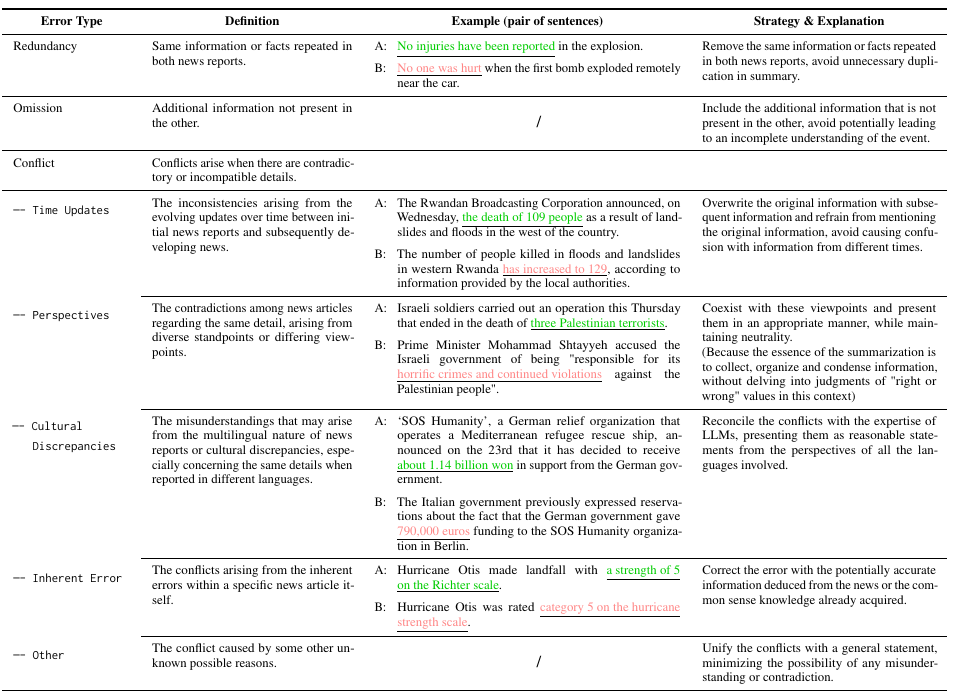}
    \caption{The protocol (English version), we formulated for MCMS task, includes definitions of all potential error types, along with corresponding real-life examples for each type and approximate resolution strategies.}
    \label{fig:error_types_en}
\end{figure*}

\subsection{Source Data Construction}\label{appendix:A-1}
\paragraph{Manual Verification} 
\textit{"high relevance does not necessarily imply that they all present the same news event"}, here is a case for distinction:
\begin{tcolorbox}[colback=yellow!15!white, colframe=white]
    \colorbox{gray!5}{\texttt{[Description]}} \hfill{\texttt{Date: 2023-10-18}} \\
    The U.S. Treasury Department announced the easing of certain oil, gas, and gold sanctions on Venezuela.
    \begin{itemize}
        \item[(1)] \colorbox{green!20!white}{\texttt{[News1]}} \hfill{\texttt{Date: 2023-10-19}} \\ After reaching an agreement, the United States lifted sanctions on Venezuelan oil and gold ... \scalebox{1.3}\greentick
        
        \item[(2)] \colorbox{red!20!white}{\texttt{[News2]}} \hfill{\texttt{Date: 2023-05-09}} \\ Maduro calls the US takeover of oil company Citgo a violation of Venezuela's sovereignty ... \scalebox{1.3}\redcross
    \end{itemize}
\end{tcolorbox}
Both the \colorbox{green!20!white}{\texttt{[News1]}} and \colorbox{red!20!white}{\texttt{[News2]}} are highly relevant in overlapping terms (e.g. Venezuela, US, ...) with the given description. Obviously \colorbox{green!20!white}{\texttt{[News1]}} is the exact news event as described in the provided description, but it's challenging for a BM25 retriever to distinguish between them.

Therefore, we incorporate a post-retrieval manual verification. 5 annotators are invited to assess the relevance of retrieved news reports based on the specified event description. Only news meeting at least one of the following criteria is retained: (1) news that describes the same event as the given query, (2) news that involves the causes and consequences of the given query event and (3) news that reflects diverse perspectives on the given query event.

\subsection{Reference Annotation Methodology}\label{appendix:A-2}
\paragraph{Key Information Split (KIS)}
In order to prevent information from becoming overly fragmented after being splitted, thereby overlooking the contextual connections, our prompt explicitly instructs the model to employ specific entity names instead of pronouns. The full request is formulated as follows:
\begin{promptbox}
	\texttt{Please split the key information of the given news article into several simple sentences in \colorbox{color1}{\{Source Language\}} and use specific entity names instead of pronouns whenever possible.} \\
	Request: \colorbox{color2}{\{Given news article $X$\}}
\end{promptbox}

\paragraph{Cross-lingual Prompting (CLP)}
The prompt is designed as:
\begin{promptbox}
	\texttt{Please act as an expert in cross-lingual understanding in \colorbox{color1}{\{Source Language\}}}. \\
	Request: \colorbox{color2}{\{Given content $X$\}} \\
    Let's understand the content in \colorbox{color3}{\{Target Language\}} sentence-by-sentence.
\end{promptbox}

\paragraph{Protocol-guided Prompting (PGP)} The full prompt for \underline{\texttt{[where]}}, \underline{\texttt{[strategies]}} and summarization process are provided in Table~\ref{tab:full_prompts}.

\section{Quality Assessment Details}\label{appendix:B}
\subsection{Variations of Precision, Recall, and F metrics} \label{appendix:B-1}
We have made slight modifications to the traditional precision, recall, and F\textsubscript{1} metrics for evaluating the performance of protocol-guided prompting with GPT-4. 
Here, we introduce the concept of False Positive Positive (FPP): predictions that do not align with the standard golden set but are still classified as correct answers after manual inspection.
The formulas are as follows:

\[Precision^* = \frac{TP + FPP}{TP + FP}\]
\[Recall^* = \frac{TP + FPP}{TP + FN}\]
\[F_{1}^* = \frac{2 * Precision * Recall}{Precision + Recall}\]

Notably, the $R^*$ value here may exceed 1, as GPT-4 has demonstrated the ability to detect additional redundancies, omissions, and conflicts overlooked by human annotators in practical testing. However, through manual verification, some of these overlooked items are also confirmed as accurate answers, thus contributing to the $R^*$ metric numerator during calculation..

\subsection{Component-wise Analysis Details}\label{appendix:B-2}
The KIS step is introduced to shorten the cumulative input length of a document set, preventing excessive length. Summarizing each document before generating an overall summary is also a commonly used method for condensing. Thus, we conducted comparative experiments on XQuAD, a multi-lingual and cross-lingual QA dataset, exploring these two methods of compression. Similarly, CLP, aiming to achieve cross-lingual alignment between source and target languages, will be compared with the method of direct translation. The experiment assesses the impact of different processing methods on LLM's comprehension in Question-Answering (QA) by applying them individually to contexts from XQuAD.

\begin{table}[t]
\resizebox{\columnwidth}{!}{%
\centering
\begin{tabular}{lrlrlr}
\hline
\textbf{Language} & \textbf{\# Docs} & \textbf{Language} & \textbf{\# Docs} & \textbf{Language} & \textbf{\# Docs}\\
\hline
Bulgarian & 230 & Swedish & 195 & Hindi & 50 \\
Italian & 254 & Hungarian & 170 & Dutch & 217 \\
Portuguese & 281 & Russian & 226 & Arabic & 202 \\
Romanian & 224 & Danish & 138 & Macedonian & 157 \\
Turkish & 211 & Ukrainian & 199 & Catalan & 79 \\
Polish & 217 & Korean & 127 & Greek & 109 \\
Finnish & 100 & Spanish & 307 & Czech & 58 \\
German & 230 & French & 281 & Hebrew & 9 \\
\cline{5-6}
Albanian & 104 & English & 312 & \textbf{Total} & \textbf{4687} \\
\hline
\end{tabular}}
\caption{Languages covered by {\datasetname} dataset, and the number of documents for each language.}\label{tab:lang_dis}
\end{table}

\begin{figure}[t]
    \centering
    \includegraphics[width=1\linewidth]{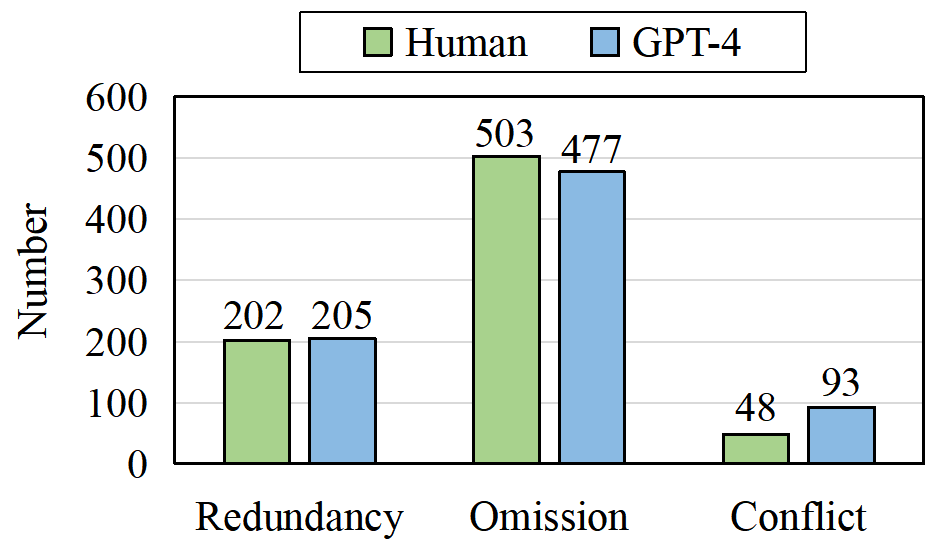}
    \caption{The comparison on the quantities of redundancies, omissions and conflicts identified by human annotators and GPT-4.}
    \label{fig:numbers}
\end{figure}

\section{Experimental Details}\label{appendix:C}
\subsection{Pipeline Overview}\label{appendix:C-1}
The detailed introductions to our pipelines are as follows:
\begin{itemize}[leftmargin=*,topsep=0pt]
\setlength{\parsep}{0pt}
\setlength{\parskip}{0pt}
\item \textbf{Translate-then-Summarize} first translates the documents into target language, then performs summarization on the translated documents.
\item \textbf{Summarize-then-Translate} first summarizes each document in the source language, then translates the summaries into target language.
\item \textbf{KIS-then-CLP} (Section~\ref{subsec:method}) first utlizes KIS step, then carries out CLP step.
\item \textbf{Direct Summarization} summarizes documents straightforwardly.
\item \textbf{Protocol-guided Prompting} (Section~\ref{subsec:method}) summarizes documents under the guidance of our protocol.
\end{itemize}

\subsection{Metrics Formulation}\label{appendix:C-2}
We evaluate the quality of summaries generated by different models and methods using following metrics:
\begin{itemize}[leftmargin=*,topsep=0pt]
\setlength{\parsep}{0pt}
\setlength{\parskip}{0pt}
\item \textbf{\textsc{ROUGE}} \citep{lin2004rouge} measures the overlap co-occurrence of n-grams between the candidate and reference summaries. We reported the $F_1$ scores for ROUGE.
\item \textbf{Red} \citep{chen-etal-2021-training} is a self-referenced metric for \textit{redundancy} evaluation. The summary itself is engaged as the reference to evaluate the degree of the semantic similarity between each summary sentences. 
The averaged semantic similarity result is used as the redundancy score.
BERTScore \citep{bert-score} is employed for similarity computation, and we use \textit{deberta-xlarge-mnli} \citep{he2021debertav3} as its backbone with its default setting for rescaling: 
\[score_{red} = \frac{\sum_i \max_{j:i \neq j} \text{Sim}(\mathbf{x}_j, \mathbf{x}_i)}{|\mathbf{X}|}, \label{equation:redund}\]
where ``$j:i \neq j$'' means we do not consider the similarity between $\mathbf{x}_i$ and itself. We use $F_1$ in as the final redundancy score. Note that $score_{red} \in [-1, 1]$ and lower is better.

\item \textbf{Normalized Inverse of Coverage (NIC)} captures \textit{Omission}, as the inverse of a coverage of key information from reference summary. We employ an NLI model \textit{t5\_xxl\_true\_nli\_mixture} \citep{honovich-etal-2022-true-evaluating} to ascertain whether crucial information from the reference is entailed in the candidate summary.
\[coverage = \frac{count(entailed)}{count(sents)}\]
\[NIC = 1 - \frac{coverage}{log(|cand|)} * 10\]
where $coverage$ represents the coverage rate of a candidate summary, $count(entailed)$ means the number of sentences from reference summary entailed in candidate summary, $count(sents)$ means the total number of sentences in reference summary and $|cand|$ means the word-level length of the candidate summary\footnote{The word-level length is calculated by nltk.}. Note that lower $NIC$ is better.

\item \textbf{Conflict Resolution Effectiveness (CRE)} metric evaluates how well a candidate summary addresses \textit{conflict}. We use GPT-3.5-turbo as a referee to assess the conflict resolution strategies presented in the candidate summary. We employ conflicts identified by GPT-4 as the standard, assessing the effectiveness of the candidate summary's handling of conflicts based on the prompts provided in Table~\ref{tab:conflict_prompt} and the result is present in a three-class classification of 1, 0, -1. In order to minimize errors resulting from conflicts update, we do not consider all conflicts identified in the summarization process for evaluation. Instead, we only select conflicts identified in the last 5 rounds of the CRS iteration process.
The Con score is calculated as follows:

\[penalty = log(count(0) + e)\]
\[score_{con} = \frac{count(1)}{count(1/-1) + \alpha * penalty} \]

where \(count(*)\) represents the counting function, \(penalty\) refers to the punitive consequence for neglecting conflict resolution. The coefficient \(\alpha\) is set to 0.2 in our work.
\end{itemize}

\subsection{Experimental implementations}\label{appendix:C-3}
\paragraph{Main experiments}
The results in Table~\ref{tab:main} only present the ROUGE-L scores, while the full ROUGE results for the main experiment can be found in Table~\ref{tab:rouge_main}.

\paragraph{Ablation study}
\begin{table}[t]
    \resizebox{\columnwidth}{!}{%
        \begin{tabular}{l|ccc}
            \hline
            \multirow{2}{*}{Schema \& Pipeline} & \multicolumn{3}{c}{GPT-3.5-turbo-16k} \\
            \cline{2-4}
            & R-1 & R-2 & R-L \\
            \noalign{\vskip 0.2ex} \hline \noalign{\vskip 0.2ex}
            \underline{Single-turn Summaization (STS)} &&&  \\
            \texttt{None + Direct}            & 33.01 & 11.02 & 17.60 \\
            \texttt{KIS-only + Direct}        & 36.37 & 12.12 & 19.60  \\
            \texttt{KIS-then-CLP + Direct}    & \textbf{41.55} & \textbf{13.76} & \textbf{21.25}  \\
            \noalign{\vskip 0.2ex}\cdashline{1-4}\noalign{\vskip 0.2ex}
            \texttt{None + Protocol}          & 35.01 & 12.16 & 19.12  \\
            \texttt{KIS-only + Protocol}      & 35.52 & 11.85 & 19.00  \\
            \texttt{KIS-then-CLP + Protocol}  & \textbf{41.58} & \textbf{14.33} & \textbf{21.17}  \\
            \noalign{\vskip 0.2ex} \hline \noalign{\vskip 0.2ex}
            \underline{Chronological Recurrent Summarization (CRS)} &&& \\
            \texttt{None + Direct}            & 34.94 & 11.08 & 18.45 \\
            \texttt{KIS-only + Direct}        & 42.12 & 14.16& 20.76 \\
            \texttt{KIS-then-CLP + Direct}    & \textbf{46.60} & \textbf{15.51} & \textbf{21.92} \\
            \noalign{\vskip 0.2ex}\cdashline{1-4}\noalign{\vskip 0.2ex}
            \texttt{None + Protocol}          & 37.40 & 12.40 & 19.47  \\
            \texttt{KIS-only + Protocol}      & 41.58 & 15.30 & 21.46  \\
            \texttt{KIS-then-CLP + Protocol}  & \textbf{47.97} & \textbf{16.41} & \textbf{22.06}  \\
            \hline
        \end{tabular}}
        \caption{
    		The full ROUGE results of ablation studies on \textit{KIS-then-CLP} stage.
    	}
	\label{tab:rouge_ablation}
    
\end{table}
The results in Table~\ref{tab:ablation} only present the ROUGE-L scores, while the full ROUGE results for the ablation study can be found in Table~\ref{tab:rouge_ablation}.

\paragraph{LLM's Scale-Effect on PGP}
As we only investigate the impact on protocol-guided prompting, in order to reduce time and computational cost, we utilize the results of the pre-steps (translate-then-summarize, summarize-then-translate, and KIS-then-CLP) from GPT-3.5-turbo-16k model. The llama2 series models are only employed in the final summarization step. Due to the 
The llama2 models are running with the default settings in this project\footnote{\url{https://github.com/notrichardren/llama-2-70b-hf-inference/blob/main/inference.py}}.

\paragraph{The Apathy towards Low-Resource Languages}
We employ an NLI model \textit{t5\_xxl\_true\_nli\_mixture} \citep{honovich-etal-2022-true-evaluating} to discern whether the sentences in the generated summary are entailed within the documents in the document set. If a sentence is entailed in a document, we consider it to be concluded from that document. 
Due to the NLI model's lack of capabilities in handling lengthy texts and multiple languages, we consider the output of the original document after undergoing KIS-then-CLP as the premise. Each sentence from the generated summary is then treated as a hypothesis. The entail score for each language is calculated as follows:
\[score_{lang} = \frac{\sum_{S} \frac{count(entailed)}{count(sents)}}{|S|}\]
\[norm\_score_{lang} = \frac{score_{lang}}{\sum\limits_{lang \in langs}score_{lang}}\]
where $S$ represents the set of generated summaries involved with the language, $count(entailed)$ means the number of sentences entailed in the document, $count(sents)$ means the total number of sentences in generated summary. $|S|$ means the number of summaries in $S$.

\section{Expenses and Compensation}\label{appendix:D}
The overall cost incurred for our reference annotation and experimental section utilizing GPT series models is approximately {\$}900.

In the manual annotation phase of post-retrieval verification (Section~\ref{subsec:source}), annotators will be compensated with {\$}0.1 for each "yes" or "no" annotation completed. We have invited a total of 5 annotators, and they have collectively annotated 18787 news articles regarding their relevance to the provided event descriptions, resulting in a total expenditure of {\$}1878.7. 
As for the construction of our protocol and the annotation process in Section~\ref{subsec:GPT4}, they are all undertaken by us paper authors without extra payment.

\begin{table*}
	\centering
	\begin{adjustbox}{width=\textwidth}
        \begin{tabular}{l|cccc|cccc|cccc}
            \hline
            \multirow{2}{*}{Schema \& Pipeline} & \multicolumn{4}{c|}{Llama-2-7b-chat-hf}  & \multicolumn{4}{c|}{Llama-2-13b-chat-hf} & \multicolumn{4}{c}{Llama-2-70b-chat-hf} \\
            \cline{2-13}
            & R-L {\uparr} & Red {\downarr} & NIC {\downarr} & Con {\uparr} & R-L {\uparr} & Red {\downarr} & NIC {\downarr} & Con {\uparr} & R-L {\uparr} & Red {\downarr} & NIC {\downarr} & Con {\uparr} \\
            \noalign{\vskip 0.2ex} \hline \noalign{\vskip 0.2ex}
            \underline{Chronological Recurrent Summarization (CRS)} &&&&&&&&& \\
            \texttt{Translate-then-Summarize + Direct}      & 19.18	& 28.82	& 84.66	& 53.01	& 17.71	& 25.66	& 88.51	& 51.21	& 18.15	& 26.26	& 81.94	& 52.47 \\
            \texttt{Translate-then-Summarize + Protocol}      & 15.52	& 40.35	& 96.04	& 45.07	& 14.88	& 25.24	& 92.95	& 50.28	& 17.49	& 31.30	& 85.93	& 56.35 \\
            \noalign{\vskip 0.2ex}\cdashline{1-13}\noalign{\vskip 0.2ex}
            \texttt{Summarize-then-Translate + Direct}                  & 19.32	& 27.03	& 93.34	& 51.98	& 17.85	& 26.44	& 88.81	& 47.74	& 17.85	& 25.38	& 85.93	& 53.81 \\
            \texttt{Summarize-then-Translate + Protocol}    & 16.06	& 37.56	& 94.92	& 50.65	& 14.66	& 27.00	& 93.51	& 51.07	& 17.19	& 35.23	& 86.09	& 58.53 \\
            \noalign{\vskip 0.2ex}\cdashline{1-13}\noalign{\vskip 0.2ex}
            \texttt{KIS-then-CLP + Direct}    & 20.67	& 29.62	& 74.53	& 51.83	& 19.33	& 26.60	& 83.44	& 50.94	& 20.39	& 29.15	& 75.03	& 53.75 \\
            \texttt{KIS-then-CLP + Protocol}                & 16.39 & 38.94	& 93.45	& 46.29	& 16.37	& 26.48	& 89.33	& 50.37	& 18.01	& 25.77	& 82.19	& 55.10 \\
            \noalign{\vskip 0.2ex} \hline \noalign{\vskip 0.2ex}
            \noalign{\vskip 0.2ex} \hline \noalign{\vskip 0.2ex}
            \texttt{Direct.Avg}                                & 19.72	& 28.49	& 80.84	& 52.27	& 18.30	& 26.23	& 86.92 & 49.96	& 18.80	& 26.93	& 80.97	& 53.34 \\
            \texttt{Protocol.Avg}                       & 15.99	& 38.95	& 94.80	& 47.34	& 15.30	& 26.24	& 91.93	& 50.57	& 17.56	& 30.77	& 85.33	& 56.66 \\
            \noalign{\vskip 0.2ex}\cdashline{1-13}\noalign{\vskip 0.2ex}
            \texttt{\(\Delta\)} = \texttt{Protocol.Avg} - \texttt{Direct.Avg}                     & - 3.73	& 10.46	& 13.96	& - 4.94	& - 3.00	& 0.01	& 5.01	& 0.61	& - 1.24	& 3.84	& 4.36 & 3.32 \\
            \hline
        \end{tabular}
    \end{adjustbox}
    \caption{
		The evaluation results for Llama2 scale-effect analysis. "\texttt{Direct}" represents direct summarization without a protocol, while "\texttt{Protocol}" signifies the summarization method utilizing the protocol-guided prompting approach.
	}
	\label{tab:llama2}
\end{table*}

\begin{table*}
	\centering
	\begin{adjustbox}{width=\textwidth}
        \begin{tabular}{l|ccc|ccc|ccc}
            \hline
            \multirow{2}{*}{Schema \& Pipeline} & \multicolumn{3}{c|}{Llama-2-7b-chat-hf}  & \multicolumn{3}{c|}{Llama-2-13b-chat-hf} & \multicolumn{3}{c}{Llama-2-70b-chat-hf} \\
            \cline{2-10}
            & R-1 & R-2 & R-L & R-1 & R-2 & R-L & R-1 & R-2 & R-L \\
            \noalign{\vskip 0.2ex} \hline \noalign{\vskip 0.2ex}
            \underline{Chronological Recurrent Summarization (CRS)} &&&&&&&&& \\
            \texttt{Translate-then-Summarize + Direct}      & 41.34	& 13.03	& 19.18	& 34.05	& 10.70	& 17.71	& 35.72	& 11.19	& 18.15 \\
            \texttt{Translate-then-Summarize + Protocol}      & 29.73	& 7.13	& 15.52	& 28.65	& 8.41	& 14.88	& 35.42	& 10.44	& 17.49 \\
            \noalign{\vskip 0.2ex}\cdashline{1-10}\noalign{\vskip 0.2ex}
            \texttt{Summarize-then-Translate + Direct}                  & 51.45	& 13.04	& 19.32	& 34.55	& 10.42	& 17.85	& 35.65	& 11.08	& 17.85 \\
            \texttt{Summarize-then-Translate + Protocol}    & 31.09	& 7.74	& 16.06	& 28.18 	& 7.68	& 14.66	& 35.07	& 9.59	& 17.19 \\
            \noalign{\vskip 0.2ex}\cdashline{1-10}\noalign{\vskip 0.2ex}
            \texttt{KIS-then-CLP + Direct}    & 43.56	& 15.19	& 20.67	& 38.31	& 13.39	& 19.33	& 40.45	& 14.39	& 20.39 \\
            \texttt{KIS-then-CLP + Protocol}                & 31.34	& 8.46	& 16.39	& 32.18	& 9.71	& 16.37	& 36.41	& 11.16	& 18.01 \\
            \noalign{\vskip 0.2ex} \hline \noalign{\vskip 0.2ex}
            \noalign{\vskip 0.2ex} \hline \noalign{\vskip 0.2ex}
            \texttt{Direct.Avg}                                & 45.45	& 13.75	& 19.72	& 35.64	& 11.50	& 18.30	& 37.27	& 12.22	& 18.80 \\
            \texttt{Protocol.Avg}                       & 34.40	& 9.27	& 16.92	& 31.16	& 9.33	& 16.05	& 36.04	& 10.85	& 17.87 \\
            \noalign{\vskip 0.2ex}\cdashline{1-10}\noalign{\vskip 0.2ex}
            \texttt{\(\Delta\)} = \texttt{Protocol.Avg} - \texttt{Direct.Avg}                     &-11.05	& -4.48	& -2.80	& -4.48	& -2.18	& -2.25	& -1.23	& -1.37	& -0.93 \\
            \hline
        \end{tabular}
    \end{adjustbox}
    \caption{
		The ROUGE results for Llama2 scale-effect analysis. "\texttt{Direct}" represents direct summarization without a protocol, while "\texttt{Protocol}" signifies the summarization method utilizing the protocol-guided prompting approach.
	}
	\label{tab:rouge_llama2}
\end{table*}

\begin{table*}
	\centering
	\begin{adjustbox}{width=\textwidth}
        \begin{tabular}{l|ccc|ccc|ccc}
            \hline
            \multirow{2}{*}{Schema \& Pipeline} & \multicolumn{3}{c|}{GPT-3.5-turbo-16k}  & \multicolumn{3}{c|}{Vicuna-7b-v1.5-16k} & \multicolumn{3}{c}{Chatglm3-6b-32k} \\
            \cline{2-10}
            & R-1 & R-2 & R-L & R-1 & R-2 & R-L & R-1 & R-2 & R-L \\
            \noalign{\vskip 0.2ex} \hline \noalign{\vskip 0.2ex}
            \underline{Single-turn Summaization (STS)} &&&&&&&&&  \\
            \texttt{Translate-then-Summarize + Direct}      & 38.63 & 12.35 & 19.86 & 37.35 & 11.20 & 18.49 & 40.32 & 11.64 & 18.98 \\
            \texttt{Summarize-then-Translate + Direct}      & 37.94 & 12.36 & 20.07 & 37.72 & 11.25 & 19.08 & 39.90 & 11.94 & 19.45 \\
            \texttt{KIS-then-CLP + Direct}                  & 41.55 & 13.76 & 21.25 & 38.53 & 11.52 & 18.96 & 40.49 & 11.59 & 19.27 \\
            \noalign{\vskip 0.2ex}\cdashline{1-10}\noalign{\vskip 0.2ex}
            \texttt{Translate-then-Summarize + Protocol}    & 37.12 & 11.95 & 19.41 & 38.14 & 11.35 & 19.07 & 38.34 & 11.50 & 18.76 \\
            \texttt{Summarize-then-Translate + Protocol}    & 36.35 & 11.62 & 19.18 & 38.23 & 11.44 & 18.69 & 38.33 & 11.23 & 18.76
            \\
            \texttt{KIS-then-CLP + Protocol}                & 41.58 & 14.33 & 21.17 & 39.57 & 11.59 & 18.82 & 39.50 & 12.04 & 19.29 \\
            \noalign{\vskip 0.2ex}\cdashline{1-10}\noalign{\vskip 0.2ex}
            \texttt{STS.Avg}                                & 38.86 & 12.73 & 20.16 & 38.26 & 11.39 & 18.85 & 39.48 & 11.66 & 19.09 \\
            \noalign{\vskip 0.2ex} \hline \noalign{\vskip 0.2ex}
            \underline{Chronological Recurrent Summarization (CRS)} &&&&&&&&& \\
            \texttt{Translate-then-Summarize + Direct}      & 41.6 & 13.09 & 20.27 & 42.89 & 13.02 & 20.14 & 44.87 & \textbf{13.87} & 20.08 \\
            \texttt{Summarize-then-Translate + Direct}      & 41.59 & 13.18 & 20.15 & 42.21 & 12.90 & 19.62 & 44.31 & 13.12 & 20.13 \\
            \texttt{KIS-then-CLP + Direct}                  & 46.6 & 15.51 & 21.92 & 44.75 & 13.74 & 20.59 & \textbf{45.57} & 13.62 & 20.12 \\
            \noalign{\vskip 0.2ex}\cdashline{1-10}\noalign{\vskip 0.2ex}
            \texttt{Translate-then-Summarize + Protocol}    & 42.78 & 14.45 & 21.14 & 44.46 & 13.77 & 20.85 & 43.90 & 13.42 & \textbf{20.48} \\
            \texttt{Summarize-then-Translate + Protocol}    & 42.68 & 14.29 & 21.24 & 43.10 & 13.09 & 20.21 & 42.52 & 12.80 & 19.60 \\
            \texttt{KIS-then-CLP + Protocol}                & \textbf{47.97} & \textbf{16.41} & \textbf{22.06} & \textbf{46.64} & \textbf{14.56} & \textbf{20.94} & 43.86 & 13.55 & 20.19 \\
            \noalign{\vskip 0.2ex}\cdashline{1-10}\noalign{\vskip 0.2ex}
            \texttt{CRS.Avg}                                & 43.87 & 14.49 & 21.13 & 44.01 & 13.51 & 20.39 & 44.17 & 13.40 & 20.10 \\
            \noalign{\vskip 0.2ex} \hline \noalign{\vskip 0.2ex}
            \noalign{\vskip 0.2ex} \hline \noalign{\vskip 0.2ex}
            \texttt{Translate-then-Summarize.Avg}           & 40.03 & 12.96 & 20.17 & 40.71 & 12.34 & 19.64 & 41.86 & 12.61 & 19.58 \\
            \texttt{Summarize-then-Translate.Avg}           & 39.64 & 12.86 & 20.16 & 40.32 & 12.17 & 19.40 & 41.27 & 12.27 & 19.49 \\
            \texttt{KIS-then-CLP.Avg}                       & \textbf{44.43} & \textbf{15.00} & \textbf{21.60} & \textbf{42.37} & \textbf{12.85} & \textbf{19.83} & \textbf{42.36} & \textbf{12.70} & \textbf{19.72} \\
            \noalign{\vskip 0.2ex}\cdashline{1-10}\noalign{\vskip 0.2ex}
            \texttt{STS + Direct.Avg}                       & \textbf{39.37} & \textbf{12.82} & \textbf{20.39} & 37.87 & 11.32 & 18.84 & \textbf{40.24} & \textbf{11.72} & \textbf{19.23} \\
            \texttt{STS + Protocol.Avg}                     & 38.35 & 12.63 & 19.92 & \textbf{38.65} & \textbf{11.46} & \textbf{18.86} & 38.72 & 11.59 & 18.94 \\
            \noalign{\vskip 0.2ex}\cdashline{1-10}\noalign{\vskip 0.2ex}
            \texttt{CRS + Direct.Avg}                       & 43.26 & 13.93 & 20.78 & 43.28 & 13.22 & 20.12 & \textbf{44.92} & \textbf{13.54} & \textbf{20.11} \\
            \texttt{CRS + Protocol.Avg}                     & \textbf{44.48} & \textbf{15.05} & \textbf{21.48} & \textbf{44.73} & \textbf{13.81} & \textbf{20.67} & 43.43 & 13.26 & 20.09 \\
            \hline
        \end{tabular}
    \end{adjustbox}
    \caption{
		The ROUGE $F_1$ scores for all configurations of schemas and pipelines on different LLMs. "\texttt{Direct}" represents direct summarization without a protocol, while "\texttt{Protocol}" signifies the summarization method utilizing the protocol-guided prompting approach. {\uparr} represents that the higher score the better and {\downarr} means the opposite.
	}
	\label{tab:rouge_main}
\end{table*}

\begin{table*}[t]
    \small
    \centering 
    \setlength\tabcolsep{3pt}
    \resizebox{\textwidth}{!}{%
        \begin{tabular}{p{\textwidth}}
        \toprule
        Prompt for \texttt{[where]} \& \texttt{[strategies]} \\
        \toprule
        From news report 1 \\
        Request: \colorbox{color2}{\{Given new1 $X_1$\}} \vspace{4pt}\newline
        From news report 2 \\
        Request: \colorbox{color2}{\{Given new2 $X_2$\}} \vspace{4pt}\newline
        The above is the key information from two different news reports about the same event. Please clearly indicate if there are any redundancies, omissions and conflicts between each numbered sentence. \\
        Definitions for "Redundancy", "Omission", "Conflict". \\
        1. Redundancy: The instances where the same information or facts are repeated in both news reports, creating unnecessary duplication. \\
        2. Omission: Omissions occur when one news report provides additional information that is not present in the other, potentially leading to an incomplete understanding of the event. \\
        3. Conflict: Conflicts arise when there are contradictory or incompatible details between the two reports, leading to confusion or doubt about the accuracy of the information. \vspace{4pt} \newline
        \colorbox{color4}{\{Response of \underline{\texttt{[where]}}\}} \vspace{0pt} \\ 
        \cdashline{1-1}\vspace{1pt} \\
        Regarding the conflicts above, kindly specify the respective conflict types and provide specific solution strategies for each conflict. \\
        - If you think the conflict arises from the updates of news events over time, please overwrite the original information with subsequent information. \\
        - If you think the conflict arises from the contradictions of diverse perspectives, please coexist with these viewpoints and present them in an appropriate manner. \\
        - If you think the conflict arises from linguistic misunderstandings or cultural discrepancies, kindly leverage your expertise to reconcile it, presenting them as reasonable statements from the perspectives of the languages involved. \\
        - If you think the conflict is caused by errors in the news report itself, please correct it with accurate information deduced from the news or the common sense knowledge you already acquired. \\
        - If you think the conflict is caused by some other unknown reasons or you can't handle the conflict with your knowledge, please use a general statement to unify them, minimizing the possibility of any misunderstanding or contradiction. \vspace{4pt} \newline
        \colorbox{color4}{\{Response of \underline{\texttt{[strategies]}}\}} \vspace{1pt} \newline \\
        \midrule
        Prompt for Summarization process \\
        \midrule
        From news report 1 \\
        Request: \colorbox{color2}{\{Given new1 $X_1$\}} \vspace{4pt} \newline
        From news report 2 \\
        Request: \colorbox{color2}{\{Given new2 $X_2$\}} \vspace{4pt} \newline
        The above are the summarized key information from two different news reports about the same event. Please follow the given [Rules] and helpful hints [where] and [strategies] to integrate the two different summaries into a overall fluent concise summary of the event. \vspace{2pt} \newline
        \text{[Rules]} \\
        1. Redundancy: Remove the same information or facts repeated in both news reports, avoid unnecessary duplication in summary. \\
        2. Omission: Include the additional information that is not present in the other, avoid potentially leading to an incomplete understanding of the event. \\
        3. Conflict: Harmonize contradictory or incompatible details in news reports in a judicious manner, there are several solution strategies for dealing with different kinds of conflicts. \\
        - If you think the conflict arises from the updates of news events over time, please overwrite the original information with subsequent information and refrain from mentioning the original information. \\
        - If you think the conflict arises from the contradictions of diverse perspectives, please coexist with these viewpoints and present them in an appropriate manner. \\
        - If you think the conflict arises from linguistic misunderstandings or cultural discrepancies, kindly leverage your expertise to reconcile it, presenting them as reasonable statements from the perspectives of the languages involved. \\
        - If you think the conflict is caused by errors in the news report itself, please correct it with accurate information deduced from the news or the common sense knowledge you already acquired. \\
        - If you think the conflict is caused by some other unknown reasons or you can't handle the conflict with your knowledge, please use a general statement to unify them, minimizing the possibility of any misunderstanding or contradiction. \vspace{4pt} \newline
        \text{[Where]} \\
        The hints guide you to identify redundancies, omissions, and conflicts. \\
        Request: \colorbox{color2}{\{{Response of \underline{\texttt{[where]}}}\}} \vspace{4pt} \newline
        \text{[Strategies]} \\
        The hints offer proposed solution strategies for each conflict that you have to strictly follow. \\
        Request: \colorbox{color2}{\{{Response of \underline{\texttt{[strategies]}}}\}} \\
        \bottomrule
        \end{tabular}
    }
    \caption{The prompts used in \underline{\texttt{[where]}}, \underline{\texttt{[strategies]}} and summarization process.}
    \label{tab:full_prompts}
\end{table*}

\begin{table*}[t]
    \small
    \centering 
    \setlength\tabcolsep{3pt}
    \resizebox{\textwidth}{!}{%
        \begin{tabular}{p{\textwidth}}
        \toprule
        Prompt for Conflict Evaluation \\
        \toprule
        You will be given a summary of multiple news articles and the summary aims to handle the conflicts between news articles. \\
        \\
        Your task is to determine whether the summary has effectively addressed the given potential conflicts. \\
        \\
        If the summary effectively addresses the conflict, answer with 1. \\
        if the summary does not involve the conflict, answer with 0. \\
        If the summary does not address the conflict, answer with -1. \\
        \\
        Conflicts: \\
        \colorbox{color2}{\{Standard conflicts identified by GPT-4\}} \\
        \\
        Summary: \\
        \colorbox{color2}{\{Candidate summary\}} \\
        \\
        For each conflict in the given conflicts, please determine whether the given summary resolves the conflict according to the rules mentioned above. Output the result in the format of a Python list, where each element in the list should be only one of the numbers 1, 0, or -1. \\
        \bottomrule
        \end{tabular}
    }
    \caption{The prompts used for conflict evaluation.}
    \label{tab:conflict_prompt}
\end{table*}

\end{document}